\documentclass[10pt,twocolumn,letterpaper]{article}

\usepackage[pagenumbers]{cvpr}

\usepackage{microtype}
\newcommand{\method}{\textit{DexGO}\xspace}

\usepackage{tabularx}
\newcolumntype{L}{>{\raggedright\arraybackslash}X}

\definecolor{cvprblue}{rgb}{0.21,0.49,0.74}
\usepackage[breaklinks,colorlinks,allcolors=cvprblue]{hyperref}
\makeatletter
\newcommand{\plainfootnote}[1]{{\let\thefootnote\relax\long\def\@makefntext##1{\noindent##1}\footnotetext{#1}}}
\makeatother

\title{DexGO: Dexterous Grasp Transition with Object Relocation}

\author{Jiaxin Li$^{*}$ \qquad Hui Zhang$^{*\dagger}$ \qquad Mark Zander \qquad Mirko Meboldt\\
\vspace{-2mm}
\\
{\normalsize
ETH Z\"urich, Switzerland}\\
{\small $^{*}$Equal contribution}\\
{\small Project page: \href{https://egoista0326.github.io/dexgo/}{https://egoista0326.github.io/dexgo/}}
}

\begin{document}
\twocolumn[{%
\renewcommand\twocolumn[1][]{#1}%
\maketitle
\begin{center}
  \centering
  \captionsetup{type=figure}
  \vspace{-5mm}
  \includegraphics[width=1.0\textwidth]{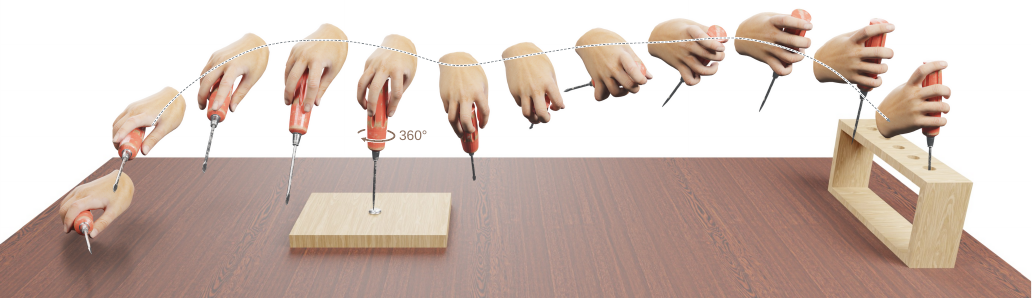}%
  \captionof{figure}{\method synthesizes dexterous manipulation motions that change the hand-object relation to a target grasp while relocating the object to a desired world pose. One policy can continuously transition between different grasp types and object poses to perform long-horizon manipulation, achieving precise and controllable motions without reference trajectories.}
  \vspace{5mm}
  \label{fig:screwdriver_transition}
\end{center}%

}]
\suppressfloats[t]
\plainfootnote{$^{\dagger}$Correspondence: huizhang@ethz.ch}
\begin{abstract}
Different tasks require an object to be held in different ways. Humans therefore often adjust their grasp while moving an object into place. Motivated by this, we study \emph{grasp transition with object relocation}: given a target grasp and a desired object world pose, the hand must transition from its initial grasp to the target grasp while relocating the object to the specified pose. This problem is challenging because the goals specify only the final state without intermediate guidance, and their complexity varies widely, from minor in-hand adjustments to grasp type changes coupled with long-distance relocations. In this work, we present \method, a reinforcement learning framework that trains a single policy through progressive task composition. Specifically, we introduce a prerequisite graph that organizes four basic capabilities and their increasingly complex combinations, allowing simpler tasks to provide intermediate guidance for learning more complex ones. We further introduce a task-wise annealed kernel reward and a nonlinear encoding of spatial observations to facilitate precise reaching across a wide range of goal distances. Experiments demonstrate that a single policy can achieve diverse target grasps and object world poses across a wide range of task complexities, producing precise and controllable motions.
\end{abstract}

\section{Introduction}
\label{sec:intro}

Manipulating an object to accomplish tasks requires more than bringing the object to a target pose, since different tasks call for different hand-object relations. For example, a hand may lift a screwdriver from the table, switch to a pinch grasp while moving it towards the screw, rotate it in hand to drive the screw in, and then change to an enveloping grasp and put the tool back into the holder (\cref{fig:screwdriver_transition}). Dexterous manipulation should therefore control both the object pose and the hand-object relation, which determines what the hand can do with the object next. As a result, the ability to generate such motions is important for different fields such as character animation and dexterous robot manipulation.

Motivated by this, we study \emph{grasp transition with object relocation}, where the hand must transform an initial grasp into a specified target grasp while moving the object to a desired world pose. The target grasp specifies the hand-object relation through the object pose relative to the hand and the finger configuration. Existing methods primarily focus on object relocation without specifying the target grasp \cite{poise2026,simtoolreal2026,play2perfect2026}. Methods that explicitly control the hand-object relation are limited to local adjustments around the initial grasp, requiring a sequence of intermediate subgoals to reach distant target grasps \cite{dexteritygen2025,dexhot2026}. Moreover, these methods do not control wrist motion, limiting their ability to adjust the grasp while bringing the object to the desired pose for subsequent manipulation. Thus, dexterous manipulation involving large grasp transitions beyond local adjustments while simultaneously relocating the object to a desired world pose remains an open problem.

This problem poses two key challenges. First, the goal specifies only the final state, leaving the intermediate motions to be discovered. For example, in \cref{fig:screwdriver_transition}, the hand has to switch the screwdriver to a pinch grasp while moving and turning it toward the screw. Such complex motions require tightly coordinated finger movements and frequent contact changes, making them difficult to discover without intermediate guidance. Second, goals vary widely in difficulty and distance, from small in-hand adjustments to grasp type changes coupled with distant relocation. Reaching these goals requires both large-scale movements toward distant targets and fine-grained adjustments for precise completion.

To address these challenges, we introduce \method, a reinforcement learning (RL) framework that trains a single policy through progressive task composition to perform grasp transitions with object relocation. Specifically, we identify four basic manipulation capabilities required for this problem: in-hand reorientation, in-hand translation, grasp type change, and object relocation, which are organized into a prerequisite graph with their combinations for progressive learning. The graph serves as a curriculum, allowing the policy to discover intermediate motions through simpler tasks and progressively learn to reach more complex and distant goals. We further introduce a task-wise annealed kernel reward that progressively narrows each task's kernel width according to its own success rate, providing long-range guidance early in training while encouraging precise goal reaching as learning progresses. Additionally, we apply a nonlinear encoding to spatial observations that magnifies small deviations while compressing large ones, enabling the policy to perceive goal distances across different manipulation scales.

We evaluate \method with diverse goals that require different manipulation capabilities and their combinations. A single policy successfully handles tasks of varying complexity, from goals that need only small changes to challenging grasp type changes with object relocation that existing methods fail to handle. We further show that different target grasps lead to distinct motions under the same initial hand-object state and desired object world pose, demonstrating the controllability of motion synthesis provided by the policy. The policy also generalizes to unseen objects, and the framework can be applied to different dexterous hands. By chaining different goals, the policy also performs long-horizon manipulation without reference trajectories. Ablation studies validate the effectiveness of our key design choices.

Our main contributions are summarized as follows:
\begin{itemize}
    \item We introduce \method, a goal-conditioned RL framework that trains a single policy to perform grasp transitions with object relocation across diverse goals, given the final target grasp and the desired object world pose.
    \item We propose a prerequisite graph that organizes tasks by their capability dependencies, enabling efficient and effective progressive learning from simple tasks to their increasingly complex combinations.
    \item We introduce a task-wise annealed kernel reward that provides guidance from distant states while enabling precise goal reaching, and propose a nonlinear spatial observation encoding for effective goal distance perception across different manipulation scales.
    \item We demonstrate that a single policy achieves diverse grasp transitions and object relocations on unseen objects, while generating distinct motions for different target grasps.
\end{itemize}

\section{Related Work}
\label{sec:related_work}
\begin{figure*}[t]
  \centering
  \includegraphics[width=\textwidth]{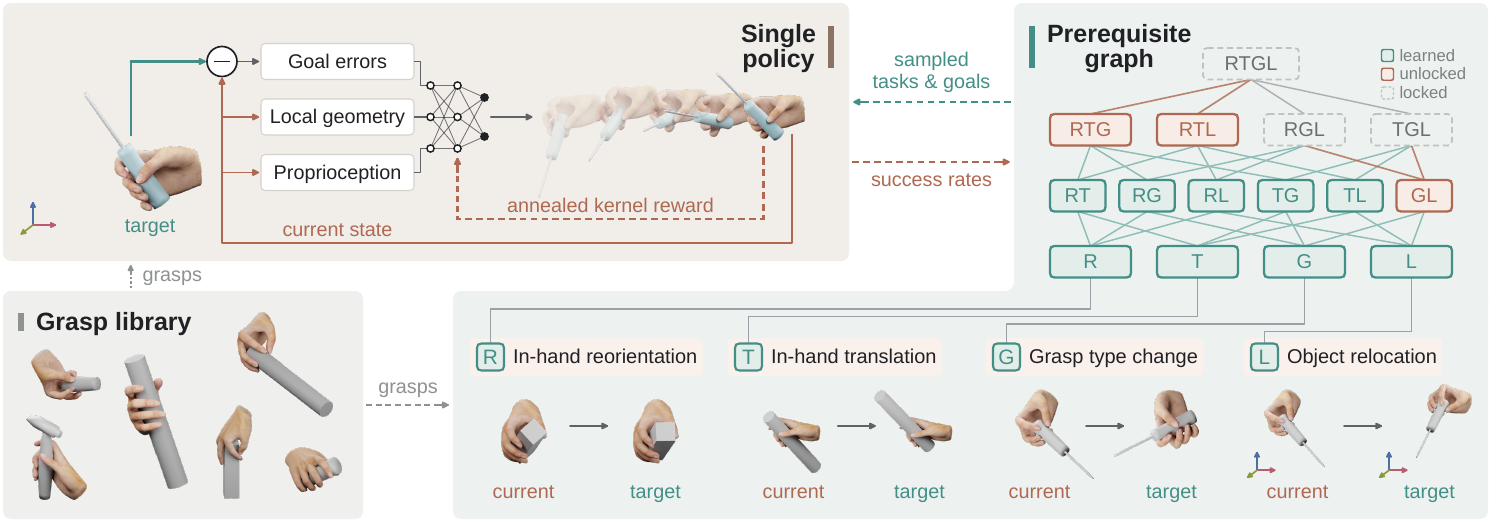}
  \caption{Overview of \method. A single policy (left) is trained on tasks that group goals by required capabilities, each with its own annealed kernel width. In the prerequisite graph (right), a task is unlocked once all its subsets with one fewer capability have been learned. A grasp library supplies both initial and target grasps (bottom left).}
  \label{fig:system_overview}
\end{figure*}

\subsection{Dexterous Manipulation}

Prior work has developed a range of dexterous manipulation capabilities, such as grasping~\cite{huang2025fungrasp,lum2024dextrahg,zhang2025RobustDexGrasp,congraspxl2026}, articulation~\cite{zhang2024artigrasp,cams2023,zhang2025bimart}, and in-hand object rotation~\cite{learningdexterity2020,inhandreorient2021,hora2023,khandate2023,poise2026,liu2026dexndm}. Beyond individual skills, some methods aim to manipulate objects toward diverse goals. For instance, SimToolReal~\cite{simtoolreal2026} learns to grasp and relocate an object to a target world pose, and Play2Perfect~\cite{play2perfect2026} further refines this ability for precise assembly. However, they do not explicitly control the final hand-object relation, which may leave the object in an unsuitable grasp for subsequent manipulation.

Some other works explicitly target hand-object relations. Planning-based approaches search for contact sequences using hand-object models~\cite{regraspplanning2018,hybridinhand2023,onlinegraspseq2023} or manually designed contact graphs~\cite{apricot2024}, often relying on object- or task-specific representations. Learning-based methods such as DexterityGen and DexHOT~\cite{dexteritygen2025,dexhot2026} train RL policies for grasp transitions toward specified hand-object relations, but focus on local adjustments around the initial grasp and do not control wrist motion, limiting their ability to relocate the object for subsequent manipulation. Unlike these methods, \method generates grasp transitions beyond local adjustments while simultaneously relocating the object to a desired world pose, given only the final goal.

Rather than generating motions directly from final goals, another line of work synthesizes hand-object interactions by following reference trajectories. Kinematic methods generate hand poses along prescribed object or wrist trajectories~\cite{manipnet2021,cams2023,manidext2025}, but geometric consistency alone does not guarantee physically feasible interactions. Physics-based methods leverage physical simulation to ensure physical plausibility. These methods typically train policies to track human motion trajectories~\cite{dextrack2025,dexplore2025,dexmachina2026,maniptrans2025} from captured datasets~\cite{fan2023arctic,zhan2024oakink2,liu2024taco}. However, such datasets are costly to collect and difficult to adapt across different hand morphologies. Unlike these approaches, \method learns to generate physically plausible hand-object motions without requiring reference trajectories.

\subsection{Cross-capability Learning}

Learning multiple capabilities within a single system has been explored through various approaches. Some methods train specialized policies separately and then combine them through policy chaining~\cite{sequentialdexterity2023}, distillation into a unified policy~\cite{unicross2026,intermimic2025,hot2025}, or modular architectures that integrate experts or primitives~\cite{mcp2019,moore2024,skillblender2025}. Other methods coordinate different capabilities across body parts, such as assigning supporting and manipulating roles to individual fingers~\cite{dexcompose2026,pei2026assembling} or employing body-part-specific discriminators~\cite{compositemotion2023}. In contrast, \method directly trains a single policy to acquire and coordinate multiple manipulation capabilities without explicit skill switching or distillation.

An alternative is to train a single policy progressively through curriculum learning, which can structure training along two dimensions: task complexity and goal precision. For task complexity, existing approaches allocate training based on learning progress~\cite{curious2019,sebn2025}, introduce control objectives in a predefined order~\cite{ulc2025}, or use manually designed prerequisite graphs~\cite{path2026}. In contrast, we define task dependencies based on the capabilities they require, progressively introducing more complex combinations as simpler tasks are learned. For goal precision, tolerance-based approaches tighten success criteria~\cite{dexpbt2023,fournier2018}, while methods like PBHC and AMS adapt exponential tracking rewards according to tracking errors~\cite{pbhc2025,ams2025}. Our task-wise annealed kernel reward instead adapts each task's kernel width independently based on its success rate, providing long-range guidance while progressively encouraging precise goal reaching.

\section{Method}
\label{sec:method}

\method trains a single policy for grasp transition with object relocation (\cref{fig:system_overview}). \Cref{sec:problem_setup} defines the problem. \Cref{sec:prerequisite_graph} introduces the basic capabilities and further organizes tasks into a prerequisite graph. \Cref{sec:goal_distance} defines a quantity that measures the deviation from the goal. \Cref{sec:reward} introduces the annealed kernel reward, which guides the policy on each task from far away to precise goal reaching. \Cref{sec:observation} describes the policy observation and action, including the nonlinear encoding of spatial observations.

\subsection{Problem Setup}
\label{sec:problem_setup}
We consider the manipulation of rigid objects with a dexterous hand. We represent poses as homogeneous transformations in $\mathrm{SE}(3)$. At time $t$, the object poses in the world frame $w$ and wrist frame $h$ are denoted by ${}_{w}G_t$ and ${}_{h}C_t$, respectively, and the finger configuration by $\mathbf{q}_t\in\mathbb{R}^n$, where $n$ is the number of actuated finger joints. The corresponding wrist pose in the world frame is ${}_{w}H_t={}_{w}G_t({}_{h}C_t)^{-1}$.

Our goal is to bring the object to a desired world pose ${}_{w}G^\star$ while reaching a target grasp specified by the relative object pose ${}_{h}C^\star$ and finger configuration $\mathbf{q}^\star$. Together, these define the goal $\mathcal K=({}_{w}G^\star,{}_{h}C^\star,\mathbf{q}^\star)$, with an implied target wrist pose ${}_{w}H^\star={}_{w}G^\star({}_{h}C^\star)^{-1}$. Given an initial grasp and this goal, the policy controls the wrist and fingers to reach this goal without a prescribed intermediate trajectory or contact sequence (\cref{fig:system_overview}, left). The initial and target grasps are sampled from a pre-defined library of feasible grasps for each object (\cref{fig:system_overview}, bottom left).

\subsection{Prerequisite Graph}
\label{sec:prerequisite_graph}
Grasp transition with object relocation requires controlling both the hand-object relation and the object world pose. A goal may require moving the object within the hand, changing the grasp type, relocating the object, or a combination of these. To capture this diversity, we identify four basic manipulation capabilities that call for different wrist-finger motion patterns: in-hand reorientation (R), in-hand translation (T), grasp type change (G), and object relocation (L). Reorientation and translation change the object pose relative to the hand while largely preserving the finger configuration. Grasp type change modifies both the finger configuration and the relative object pose, while object relocation changes the object pose in the world frame.

Different goals can be characterized by different capabilities or their combinations, resulting in varying levels of task complexity. Learning complex combinations like \cref{fig:screwdriver_transition} directly is challenging, as the policy must acquire and coordinate multiple capabilities simultaneously. However, capabilities learned from simpler tasks can facilitate learning their more complex combinations~\cite{bengio2009curriculum,lee2024shaping}. We therefore group goals into tasks according to their required capabilities and organize these tasks into a prerequisite graph (\cref{fig:system_overview}, right). Each node represents a nonempty subset of the four capabilities, resulting in 15 tasks. A task depends on all its immediate subsets, each requiring one fewer capability.

We use this graph to progressively introduce tasks during training. Initially, only the four single-capability tasks are available. A combination task is unlocked once all its prerequisites have been learned, as determined by the task-wise kernel width (\cref{sec:reward}). Among unlocked tasks, we adapt their sampling probabilities according to their success rates. Specifically, each task $i$ is assigned a weight $w_i=\mathbb{I}_i(1-p_i)$, where $\mathbb{I}_i$ indicates whether the task is unlocked and $p_i$ is its success rate under a fixed tolerance (\cref{sec:goal_distance}). Each environment samples task $i$ with probability $w_i/\sum_j w_j$, prioritizing tasks with lower success rates. In this way, the policy progressively learns more complex capability combinations while focusing training on tasks that remain challenging.

\subsection{Goal Distance}
\label{sec:goal_distance}
To measure how far the current state is from the goal, we define a scalar goal distance that combines the wrist pose error, the hand-object error, and the finger error:
\begin{equation}
  E_t=\bigl\|\bigl(\psi(\epsilon_{p,t}^H),\,\psi(\epsilon_{r,t}^H),\,\epsilon_{p,t}^C,\,\epsilon_{r,t}^C,\,P_t\bigr)\bigr\|_2.
  \label{eq:goal_distance}
\end{equation}
Here, $\epsilon_{p,t}^X=d_{p,t}^X/s_p^X$ and $\epsilon_{r,t}^X=d_{r,t}^X/s_r^X$ are the normalized position and rotation errors, where $d_{p,t}^X$ and $d_{r,t}^X$ measure the deviations from the corresponding target pose. For $X=H$, they describe the wrist pose in the world frame; for $X=C$, they describe the object pose relative to the hand. The normalization scales $s_p^X$ and $s_r^X$ are chosen based on the evaluation tolerances. The finger error $P_t={\tfrac{1}{n}\sum_{j}([|q_{t,j}-q_j^\star|-\delta_q]_+/s_q)^2}$ measures the deviation from $\mathbf{q}^\star$, where $s_q$ is a joint tolerance, $\delta_q$ is a small dead band, and $[x]_+=\max(x,0)$. Together, these terms capture errors in both the target grasp and the desired object world pose, since ${}_{w}G={}_{w}H\,{}_{h}C$.

During large object relocations, the wrist pose error can be much larger than the hand-object error and dominate $E_t$, diminishing the contribution of improvements in the hand-object relation to the kernel reward (\cref{sec:reward}). To prevent this, we compress the wrist position and rotation errors using
\begin{equation}
  \psi(\epsilon)=
  \begin{cases}
    \epsilon, & \epsilon\le\kappa,\\
    \kappa\bigl(1+\log(\epsilon/\kappa)\bigr), & \epsilon>\kappa,
  \end{cases}
  \label{eq:wrist_compression}
\end{equation}
where $\kappa$ is a threshold set separately for position and rotation. The function leaves small errors unchanged and grows logarithmically beyond $\kappa$, preserving sensitivity near the goal while preventing large wrist errors from overwhelming improvements in the hand-object relation.

\begin{figure}[t]
  \centering
  \includegraphics[width=\linewidth]{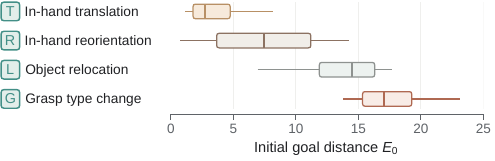}
  \caption{Distribution of initial goal distances $E_0$ across the four single-capability tasks. Boxes indicate the interquartile ranges, whiskers the 5th--95th percentiles, and ticks the medians. In-hand translation has the smallest initial goal distances, while grasp type change has the largest.}
  \vspace{-1mm}
  \label{fig:e0}
\end{figure}

A goal is considered complete when $E_t$ remains below a tolerance for a short duration, and the success rate is the fraction of goals completed under this criterion. We denote the initial goal distance of each episode by $E_0$ (\cref{fig:e0}).

\subsection{Task-Wise Annealed Kernel Reward}
\label{sec:reward}
We use a reward $r_t=r_t^{\mathrm{kernel}}+r_t^{\mathrm{drop}}+r_t^{\mathrm{reg}}$, consisting of a kernel term, a drop penalty, and regularization terms. Dropping the object terminates the episode with a penalty $r_t^{\mathrm{drop}}$. The regularization term encourages stable hand-object contact and smooth, low-effort motions: $r_t^{\mathrm{reg}}=\lambda_c c_t-\lambda_\tau\|\boldsymbol{\tau}_t\|_2^2-\lambda_a\|\Delta^2\mathbf a_t\|_2^2$, where $c_t$ measures fingertip contact coverage, $\boldsymbol{\tau}_t$ denotes finger joint torques, and $\Delta^2\mathbf a_t=\mathbf a_t-2\mathbf a_{t-1}+\mathbf a_{t-2}$ is the second-order action difference.

The kernel term measures progress toward the goal using the goal distance $E_t$ (\cref{sec:goal_distance}):
\begin{equation}
  r_t^{\mathrm{kernel}}=\exp\!\left(-\frac{E_t^2}{2\sigma^2}\right),
\end{equation}
where $\sigma$ is the kernel width that determines the range of goal distances over which the reward provides useful guidance. A small $\sigma$ encourages precise goal reaching but produces nearly zero reward for goals with large initial goal distances $E_0$, making them difficult to learn. A large $\sigma$, on the other hand, provides guidance from farther away but reduces reward sensitivity near the goal. We therefore gradually reduce $\sigma$ as the policy improves, moving from coarse guidance to precise goal reaching. During training, each task's $\sigma$ is reduced to the next smaller value once its success rate exceeds $\eta$ using the current $\sigma$ as the tolerance.

Tasks requiring different capabilities differ in the wrist-finger motion patterns they involve and in their initial goal distances $E_0$, and therefore in how difficult they are to learn. Easier tasks, such as in-hand translation, typically start closer to their goals and achieve higher success rates earlier in training, while more challenging tasks, such as grasp type change, start farther from their goals and learn more slowly (\cref{fig:e0}). To prevent successes on easier tasks from causing the kernel width to shrink prematurely for harder tasks, we assign each task (\cref{sec:prerequisite_graph}) its own kernel width, which is annealed independently based on its success rate. Each width thus narrows at its own pace. On tasks it already solves, the policy is pushed toward precise goal reaching, while tasks it has yet to solve keep long-range guidance. Because these widths adapt automatically, the same reward serves all tasks without setting or tuning a width for any individual task. A task is considered learned once its kernel width falls below $\sigma_{\mathrm{learned}}$, which determines when its dependent tasks are unlocked in the prerequisite graph.

\subsection{Observation and Action}
\label{sec:observation}
\noindent\textbf{Observation.} The actor receives three groups of observations: goal errors, local geometry, and proprioception (\cref{fig:system_overview}):
\begin{equation}
  \mathbf{o}_t=
\bigl[
\underbrace{\vphantom{\mathbf{q}_t,\mathbf{e}_t^{\mathrm{pd}},{}_{h}\hat{\mathbf{g}}_t}{}_{h}\mathbf{e}_t^G,{}_{h}\mathbf{e}_t^C,{}_{h}\mathbf{e}_t^{H},\mathbf{e}_t^q}_{\text{\strut goal errors}},
\underbrace{\vphantom{\mathbf{q}_t,\mathbf{e}_t^{\mathrm{pd}},{}_{h}\hat{\mathbf{g}}_t}{}_{h}\mathbf{v}_t}_{\text{\strut local geometry}},
\underbrace{\mathbf{q}_t,\mathbf{e}_t^{\mathrm{pd}},{}_{h}\hat{\mathbf{g}}_t}_{\text{\strut proprioception}}
\bigr].
  \label{eq:actor_observation}
\end{equation}
The goal errors include deviations in the object world pose, the hand-object relative pose, the wrist world pose, and the finger configuration. The three pose errors ${}_{h}\mathbf{e}_t^G,{}_{h}\mathbf{e}_t^C,{}_{h}\mathbf{e}_t^{H}\in\mathbb{R}^6$ each contain three position and three rotation components, while $\mathbf{e}_t^q\in\mathbb{R}^n$ measures the deviation from the target finger configuration $\mathbf{q}^\star$. Inspired by~\cite{unicross2026}, the local geometry ${}_{h}\mathbf{v}_t\in\mathbb{R}^{3M}$ consists of vectors from the $M$ hand links to their nearest sampled object surface points. Proprioception includes the finger configuration $\mathbf{q}_t$, actuator tracking errors $\mathbf{e}_t^{\mathrm{pd}}\in\mathbb{R}^n$, and gravity direction ${}_{h}\hat{\mathbf{g}}_t\in\mathbb{R}^3$. The critic additionally observes privileged simulation information, including contact forces.

\noindent\textbf{Nonlinear Encoding.} We express position errors, rotation errors, and local geometry in the current wrist frame, in which the target grasp is defined. For a rotation error ${}_{h}R_{\mathrm{err}}$, we use the rotation vector ${}_{h}\boldsymbol{\phi}=\operatorname{Log}({}_{h}R_{\mathrm{err}})$, whose direction and magnitude give the rotation axis and angle. The magnitudes of these quantities vary widely within an episode. With a linear scaling that covers this range, small deviations near the goal would be hard for the policy to perceive. We therefore apply a nonlinear encoding to a spatial vector ${}_{h}\mathbf{x}$:
\begin{equation}
  f({}_{h}\mathbf{x})=
\frac{{}_{h}\mathbf{x}}{\|{}_{h}\mathbf{x}\|}
\frac{\log\!\left(1+\min(\|{}_{h}\mathbf{x}\|,d_{\max})/d_0\right)}
{\log(1+d_{\max}/d_0)}.
  \label{eq:goal_encoding}
\end{equation}
The positive scales $d_0$ and $d_{\max}$ control sensitivity and saturation, respectively. This encoding preserves direction while magnifying small deviations and compressing large ones. We apply $f$ to each position error, rotation vector, and local geometry vector, with the scales set according to the range of each quantity.

\noindent\textbf{Action.} The policy outputs incremental actions $\mathbf{a}_t=[{}_{h}\mathbf{a}_t^p,{}_{h}\mathbf{a}_t^r,\mathbf{a}_t^q]\in\mathbb{R}^{3+3+n}$, specifying wrist translation and wrist rotation in the current wrist frame, and finger joint commands. We clip the output elementwise, $\tilde{\mathbf{a}}_t=\operatorname{clip}(\mathbf{a}_t,-1,1)$, and apply positive scales $\alpha_p,\alpha_r,\alpha_q$:
\begin{equation}
  {}_{h}\Delta\mathbf{p}_t=\alpha_p\,{}_{h}\tilde{\mathbf{a}}_t^p,
\quad
{}_{h}\Delta\boldsymbol{\theta}_t=\alpha_r\,{}_{h}\tilde{\mathbf{a}}_t^r,
\quad
\Delta\mathbf{q}_t=\alpha_q\tilde{\mathbf{a}}_t^q.
  \label{eq:action_increments}
\end{equation}
The wrist increments are applied relative to the current measured wrist pose~\cite{maniskill22023,factory2022}, while the finger increments update the previous joint commands within the joint limits. The resulting wrist and finger targets are then transferred to simulated joint torques through PD controllers.

\section{Experiments}
\label{sec:experiments}

\Cref{sec:experimental_setup} describes the experimental setup. \Cref{sec:results} evaluates how precisely one policy reaches goals that combine object relocation with different changes in the hand-object relation. \Cref{sec:qualitative} shows motions for different target grasps and long-horizon manipulation. \Cref{sec:generalization} tests generalization to everyday objects and another hand. \Cref{sec:ablation_study} ablates the main components of \method.

\subsection{Experimental Setup}
\label{sec:experimental_setup}

\noindent\textbf{Training Setup.} We train a single policy using PPO~\cite{schulman2017ppo} in Isaac Lab~\cite{mittal2025isaaclab}, with 8{,}000 parallel environments. We use MANO~\cite{romero2017mano} to construct a simulated hand with 26 degrees of freedom (DoF), including 20 finger joints and a 6-DoF wrist. The policy runs at 20\,Hz, with training episodes lasting up to 600 control steps. Training takes approximately 48\,h on a single NVIDIA RTX 4090 GPU.

During training, we randomize finger PD gains, add noise to finger joint observations, and apply random external force pulses to the object. For task-wise kernel annealing, each task starts with a width of $\sigma=10$, which is reduced whenever its success rate at the current width exceeds $\eta=0.4$. A task is considered learned once its width falls below $\sigma_{\mathrm{learned}}=2$. The kernel width schedule and other hyperparameters are provided in the supplementary material.

\noindent\textbf{Objects and Grasp Library.} We train and evaluate on randomly generated elongated objects, as many everyday objects and tool handles have such shapes~\cite{feix2014analysis}. Each object has a long axis of 10-32\,cm and two short axes of 3-5\,cm, with round, square, or hexagonal cross sections. Some are also bulged or tapered along their length. We use 123 objects for training and 46 unseen objects for evaluation. Object masses range from 0.15 to 0.30\,kg, friction coefficients from 0.5 to 1.0, and centers of mass are sampled uniformly between one quarter and three quarters of their length.

Humans use different grasp types depending on how they intend to manipulate an object. For elongated objects, a pinch grasp typically holds an end with the fingertips, while an enveloping grasp wraps the fingers and palm around the object~\cite{feix2014analysis}. To cover these distinct grasp types across all tasks, we construct a library of pinch and enveloping grasps for each object. We generate them using separate grasping policies trained with the framework of~\cite{unicross2026}, varying the grasp location and hand approach direction. Only grasps that pass lifting and contact checks are retained. The resulting MANO grasp library contains 46{,}754 enveloping grasps and 132{,}200 pinch grasps.

\noindent\textbf{Goal Generation.} We generate goals using the four capabilities defined in \cref{sec:prerequisite_graph}. Grasp type change (G) switches between pinch and enveloping grasps, involving substantial changes in the hand-object relation. On average, these goals rotate the object by approximately $90^\circ$ relative to the hand and change finger joint angles by $30^\circ$, with the largest joint changes exceeding $70^\circ$. They also have the largest initial goal distances among the four single-capability tasks (\cref{fig:e0}). We pair pinch and enveloping grasps near the same end of the object to avoid introducing additional axial sliding or rolling. For the remaining capabilities, we uniformly sample in-hand translation (T) along the object's long axis from 1 to 10\,cm, in-hand reorientation (R) about this axis up to $150^\circ$, and object relocation (L) up to 60\,cm and $120^\circ$ in world position and orientation, respectively, with random directions.

Combination goals apply the changes of multiple capabilities, each sampled from the same range as its corresponding single-capability task. For example, a TL goal combines axial in-hand translation with object relocation. Each episode starts with a grasp sampled from the library and a wrist orientation sampled uniformly over 3D rotations.

\noindent\textbf{Evaluation Protocol.} We evaluate on 46 unseen test objects, with all goals requiring object relocation (L). Each evaluation episode lasts up to 1200 control steps for all methods. We group the goals according to their required manipulation capabilities: same-type goals (TL, RL, RTL) preserve the grasp type, type-change goals (GL, TGL, RGL) require a grasp type change, and full goals (RTGL) combine all four capabilities. An episode succeeds if the object remains grasped without dropping, and all the following criteria are satisfied for five consecutive control steps:
\begin{enumerate}[nosep,leftmargin=*]
  \item The object world pose and hand-object relative pose are within 5\,mm and $5^\circ$ of their targets.
  \item The finger error $P_t\leq0.5$ (approximately $12^\circ$ average joint deviation), with at least two fingers in contact.
  \item The object's linear and angular velocities are below 2\,cm/s and 0.4\,rad/s, respectively.
\end{enumerate}

\subsection{Quantitative Evaluation}
\label{sec:results}
We first evaluate our policy under the default criteria and compare with DexterityGen (DexGen)~\cite{dexteritygen2025}, the closest prior RL method for grasp transitions. Since its goal also specifies the hand-object relation, we adapt its RL grasp-transition stage to our setting, with the same hand, objects, goals, and training budget. For a more comprehensive comparison, we also evaluate under relaxed criteria, increasing the pose tolerances to 10\,mm and $10^\circ$ and the finger error threshold to $P_t\leq1$ (approximately $15^\circ$ average joint deviation).

\Cref{tab:baselines} reports the success rates of both methods on the test objects. For relatively simple goals without grasp type change, DexGen achieves reasonable success under the relaxed criteria (81.2\%), but drops sharply to 44.0\% under the default criteria. In contrast, our policy achieves high success rates and consistently outperforms DexGen under both criteria, with only a small drop under the more stringent default criteria. We attribute this improvement in precision to our annealed kernel reward, which maintains reward sensitivity to small goal distances as the policy approaches the target, encouraging precise goal reaching.

For more challenging goals involving grasp type changes, DexGen achieves less than 1\% success. Examining its rollouts, we find that the finger joints can roughly move toward the target finger configuration, while the relative hand-object pose changes very little. This suggests that DexGen can perform local finger adjustments but struggles with the larger changes in hand-object relative pose that are required for grasp type changes. In contrast, our policy achieves 67.8\% success on type-change goals and 60.5\% on full goals even under the default criteria, demonstrating its ability to coordinate substantial grasp transitions with object relocation.

\begin{table}[t]
  \centering
  \caption{Success rates (\%) under the default and relaxed criteria.}
  \label{tab:baselines}
  \resizebox{\columnwidth}{!}{%
  \begin{tabular}{l|cccc|cccc|c}
    \toprule
    & \multicolumn{4}{c|}{Same type}
      & \multicolumn{4}{c|}{Type change}
      & \multicolumn{1}{c}{Full} \\
    \cmidrule(lr){2-5}\cmidrule(lr){6-9}\cmidrule(lr){10-10}
    Method & TL & RL & RTL & Avg. & GL & TGL & RGL & Avg. & RTGL \\
    \midrule
    \multicolumn{10}{c}{\textit{Default criteria}} \\
    \midrule
    DexGen & 47.5 & 39.3 & 45.3 & 44.0 & 0.1 & 0.1 & 0.3 & 0.2 & 0.1 \\
    Ours & \textbf{96.0} & \textbf{87.9} & \textbf{92.3} & \textbf{92.1} & \textbf{73.1} & \textbf{65.3} & \textbf{64.9} & \textbf{67.8} & \textbf{60.5} \\
    \midrule
    \multicolumn{10}{c}{\textit{Relaxed criteria}} \\
    \midrule
    DexGen & 90.9 & 66.8 & 85.9 & 81.2 & 0.3 & 0.7 & 0.6 & 0.5 & 0.4 \\
    Ours & \textbf{98.0} & \textbf{92.1} & \textbf{94.1} & \textbf{94.7} & \textbf{76.3} & \textbf{68.2} & \textbf{67.1} & \textbf{70.6} & \textbf{63.1} \\
    \bottomrule
  \end{tabular}%
  }
\end{table}

\subsection{Qualitative Results}
\label{sec:qualitative}

\begin{figure}[t]
  \centering
  \setlength{\tabcolsep}{0pt}
  \begin{tabular}{@{}l@{\hspace{0.02\columnwidth}}l@{}}
    \includegraphics[trim=21pt 62pt 15pt 43pt,clip,width=0.4900\columnwidth]{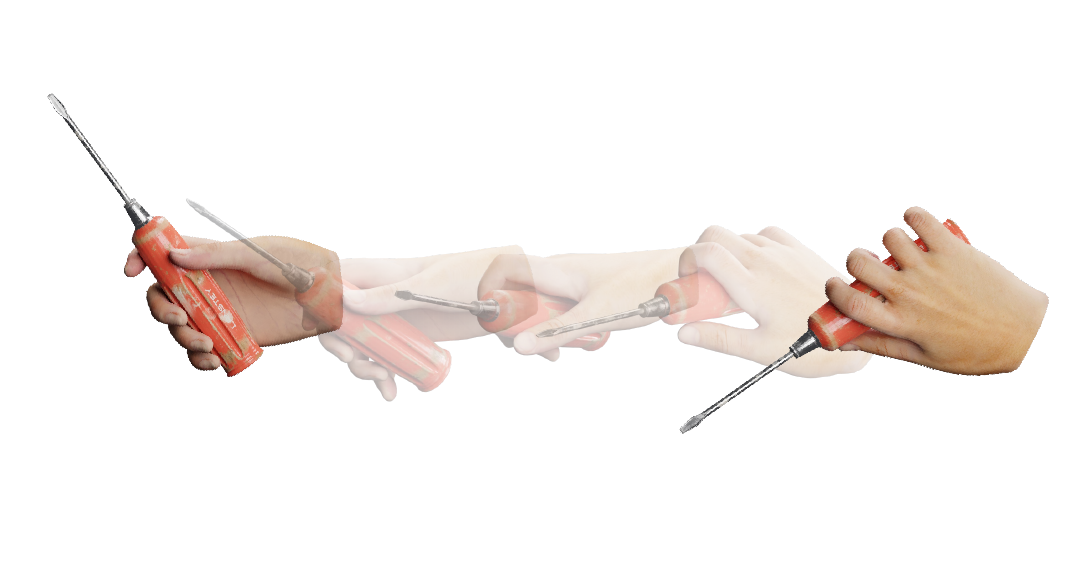} &
    \includegraphics[trim=3pt 2pt 3pt 3pt,clip,width=0.4758\columnwidth]{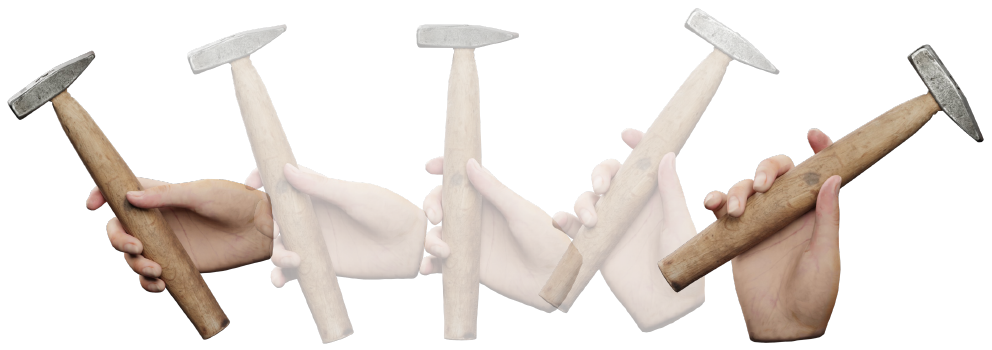}\\[-1pt]
    \makebox[0.49\columnwidth]{\footnotesize (a) Screwdriver, L} & \makebox[0.49\columnwidth]{\footnotesize (c) Hammer, L}\\[3pt]
    \includegraphics[trim=21pt 22pt 37pt 42pt,clip,width=0.4698\columnwidth]{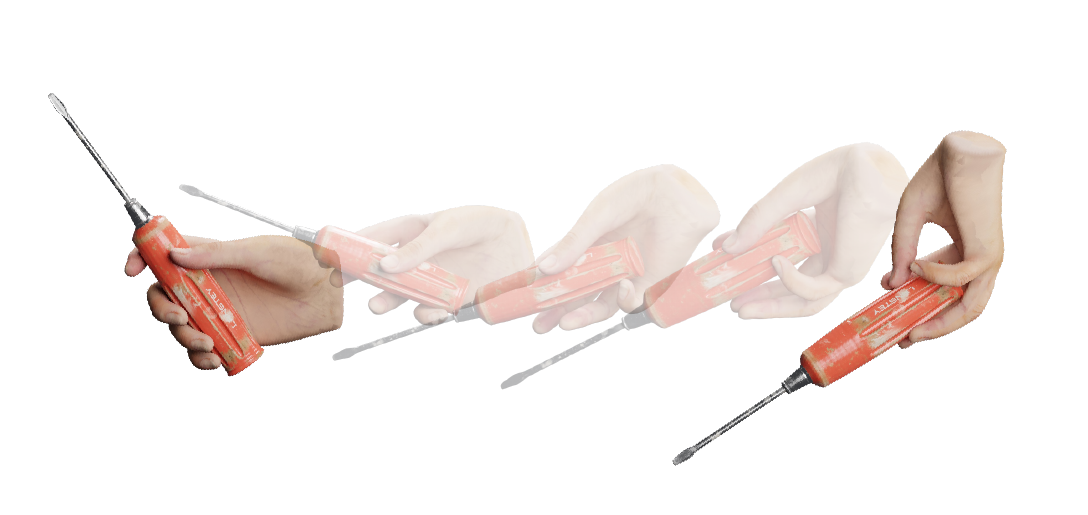} &
    \includegraphics[trim=2pt 2pt 3pt 3pt,clip,width=0.4779\columnwidth]{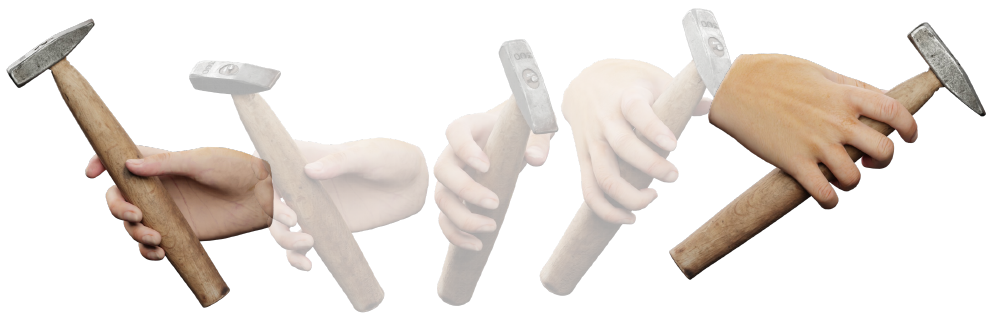}\\[-1pt]
    \makebox[0.49\columnwidth]{\footnotesize (b) Screwdriver, GL} & \makebox[0.49\columnwidth]{\footnotesize (d) Hammer, RTL}
  \end{tabular}
  \caption{Distinct motions generated by a single policy for different target grasps. The screwdriver pair (a, b) and hammer pair (c, d) each share the same initial grasp, initial object pose, and desired object world pose, but differ in their target grasps.
  }
  \label{fig:grasp_controllability}
\end{figure}

\begin{figure}[t]
  \centering
  \includegraphics[width=\columnwidth]{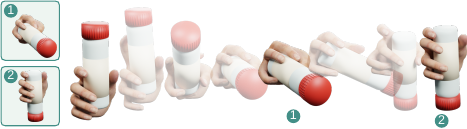}
  \caption{Consecutive grasp type changes on a spice bottle guided by a pair of goals on the left. Numbers mark the frames in which the corresponding goal is reached.
  }
  \label{fig:long_horizon}
\end{figure}

\noindent\textbf{Grasp Controllability.} Our policy demonstrates precise control over both the object world pose and the hand-object relation. To illustrate this, \cref{fig:grasp_controllability} shows two common tools unseen during training: a screwdriver and a hammer. For each object, we specify two different target grasps with the same initial hand-object state and desired object world pose. The policy generates distinct motions that accurately reach the corresponding target grasps while bringing the object to the same desired world pose, demonstrating its ability to control the final grasp independently of the world pose.

\noindent\textbf{Long-Horizon Manipulation.} By specifying a sequence of discrete goals with different target grasps and desired object world poses, our policy can generate long-horizon motions that combine object relocation with continuous hand-object relation changes. \Cref{fig:screwdriver_transition} illustrates a screwdriver example starting just after the tool is lifted from the table. The policy first moves the screwdriver toward the screw while transitioning from an enveloping to a pinch grasp, then rotates it about its long axis in hand to drive the screw. Finally, it transitions to another enveloping grasp and relocates the screwdriver above its holder for placement. \Cref{fig:long_horizon} shows another example with two goals: the policy first transitions from an enveloping to a pinch grasp, then transitions back to another enveloping grasp with a different orientation.

\subsection{Generalization}
\label{sec:generalization}

\begin{figure}[t]
  \centering
  \includegraphics[width=\columnwidth]{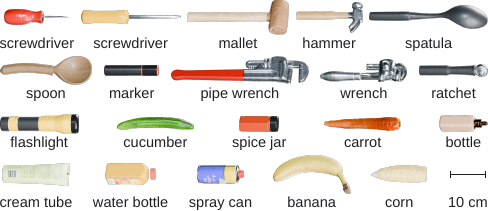}
  \caption{Everyday objects used for generalization evaluation.}
  \label{fig:everyday_objects}
\end{figure}

\noindent\textbf{Everyday Objects.} To evaluate generalization to object geometries beyond the training distribution, we construct a test set of 20 everyday objects, including ten tools and ten household items (\cref{fig:everyday_objects}). We evaluate the policy on goals relevant to the everyday use of each object, with detailed goal settings provided in the supplementary material. As shown in \cref{tab:everyday_objects}, the policy achieves success rates of 88.9\%, 60.1\%, and 48.9\% on same-type, type-change, and full goals, respectively, without further training. Despite some performance degradation due to substantial geometric differences from the training objects, such as the flat handle of a wrench and the curved axis of a banana, our policy achieves overall promising performance on these unseen everyday objects, demonstrating generalization to novel object geometries. \Cref{fig:grasp_controllability,fig:long_horizon} provide some examples of the generated motions on unseen everyday objects, with more shown in the supplementary video.

\begin{table}[t]
  \centering
  \caption{Success rates (\%) on unseen everyday objects.}
  \label{tab:everyday_objects}
  \resizebox{\columnwidth}{!}{%
  \begin{tabular}{l|cccc|cccc|c}
    \toprule
    & \multicolumn{4}{c|}{Same type}
      & \multicolumn{4}{c|}{Type change}
      & \multicolumn{1}{c}{Full} \\
    \cmidrule(lr){2-5}\cmidrule(lr){6-9}\cmidrule(lr){10-10}
    Method & TL & RL & RTL & Avg. & GL & TGL & RGL & Avg. & RTGL \\
    \midrule
    Ours & 92.8 & 85.2 & 88.6 & 88.9 & 67.0 & 53.3 & 60.1 & 60.1 & 48.9 \\
    \bottomrule
  \end{tabular}%
  }
\end{table}

\begin{table}[t]
  \centering
  \caption{Success rates (\%) on different hands.}
  \label{tab:hands}
  \resizebox{\columnwidth}{!}{%
  \begin{tabular}{l|cccc|cccc|c}
    \toprule
    & \multicolumn{4}{c|}{Same type}
      & \multicolumn{4}{c|}{Type change}
      & \multicolumn{1}{c}{Full} \\
    \cmidrule(lr){2-5}\cmidrule(lr){6-9}\cmidrule(lr){10-10}
    Hand & TL & RL & RTL & Avg. & GL & TGL & RGL & Avg. & RTGL \\
    \midrule
    MANO   & 96.0 & 87.9 & 92.3 & 92.1 & 73.1 & 65.3 & 64.9 & 67.8 & 60.5 \\
    Sharpa & 93.8 & 81.9 & 90.0 & 88.5 & 69.9 & 69.9 & 58.9 & 66.2 & 59.5 \\
    \bottomrule
  \end{tabular}%
  }
\end{table}

\noindent\textbf{Other Hands.} To evaluate the cross-hand applicability of our method involving the prerequisite graph, annealed kernel reward, and nonlinear observation encoding, we also train \method on the Sharpa Wave hand~\cite{sharpawave2025}, which has five fingers and 28 DoF (22 for the fingers and 6 for the wrist). We construct its grasp library using the same procedure and adapt the observation dimensions accordingly, while keeping the objects, goal ranges, training budget, reward formulation, and all other settings unchanged without hand-specific tuning. We found that Sharpa learns grasp type changes slower than MANO, likely because its approximately 30\% longer fingers tend to clamp thin objects at the fingertips before the objects reach the palm during enveloping grasp formation. Nevertheless, our framework achieves highly comparable overall performance on Sharpa (\cref{tab:hands}), demonstrating its applicability across different hand embodiments.

\subsection{Ablation Study}
\label{sec:ablation_study}

We conduct ablation studies to evaluate the contributions of kernel annealing, progressive task composition, and nonlinear observation encoding, with the results shown in \cref{tab:ablation}. \emph{Fixed width} replaces annealing with $\sigma=2$, 6, or 10 while retaining the prerequisite graph; a task is considered learned when its success rate at $\sigma_{\mathrm{learned}}$ exceeds $\eta$. \emph{W/o unlocking} trains all capability combinations from the start while retaining task-wise annealing. \emph{W/o graph} removes both task grouping and unlocking, using a single annealed kernel width for all goals. \emph{Linear encoding} replaces our nonlinear spatial encoding (\cref{eq:goal_encoding}) with a linear representation. We report success under both default (Def.) and relaxed (Rel.) criteria (\cref{sec:results}) for precision comparison.

\begin{table}[t]
  \centering
  \caption{Ablation study.}
  \label{tab:ablation}
  \resizebox{1.0\columnwidth}{!}{%
  \begin{tabular}{l|cc|cc|cc}
    \toprule
    & \multicolumn{2}{c|}{Same type}
      & \multicolumn{2}{c|}{Type change}
      & \multicolumn{2}{c}{Full} \\
    \cmidrule(lr){2-3}\cmidrule(lr){4-5}\cmidrule(lr){6-7}
    Setting & Def. & Rel. & Def. & Rel. & Def. & Rel. \\
    \midrule
    Fixed width ($\sigma=2$)    & 83.3 & 93.3 & 4.0 & 6.0 & 1.4 & 2.0 \\
    Fixed width ($\sigma=6$)    & 68.4 & 89.5 & 34.8 & 50.4 & 28.2 & 37.0 \\
    Fixed width ($\sigma=10$)   & 52.9 & 81.8 & 13.3 & 21.0 & 9.6 & 15.2 \\
    \midrule
    W/o unlocking       & 89.9 & 93.8 & 61.0 & 66.1 & 50.5 & 54.5 \\
    W/o graph           & 88.6 & 93.6 & 38.6 & 39.9 & 35.5 & 36.5 \\
    \midrule
    Linear encoding     & 84.8 & 90.7 & 59.5 & 66.6 & 48.2 & 53.8 \\
    \midrule
    Ours                & \textbf{92.1} & \textbf{94.7} & \textbf{67.8} & \textbf{70.6} & \textbf{60.5} & \textbf{63.1} \\
    \bottomrule
  \end{tabular}%
  }
\end{table}

\begin{figure}[t]
  \centering
  \includegraphics[width=0.8\columnwidth]{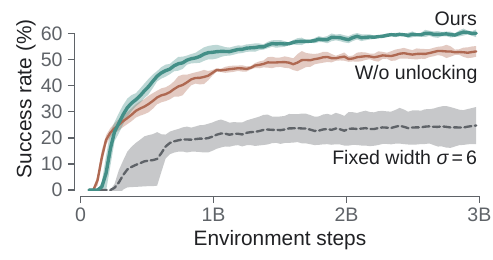}
  \caption{Success rates on GL goals (grasp type change and object relocation) measured during training under the default criteria. Success is zero before task unlocking. Curves are smoothed over 500 policy updates. Each line shows the mean across three seeds, with the shaded area indicating the standard deviation.}
  \label{fig:gl_training}
\end{figure}

\noindent\textbf{Kernel Annealing.} A narrow fixed kernel ($\sigma=2$) works relatively well for simpler tasks with small initial goal distances $E_0$, but provides little guidance for distant goals involving grasp type changes (\cref{fig:e0}), and thus fails to learn these challenging tasks. Wider fixed kernels ($\sigma=6$ and 10) improve long-range guidance but reduce near-goal sensitivity, leading to less precise goal reaching and thus substantially lower success, especially under the default criteria. In contrast, our annealed kernel provides both long-range guidance early in training while progressively increasing sensitivity to fine adjustments near the goal, achieving 92.1\% success on same-type goals and 67.8\% on type-change goals under the default criteria. Moreover, it accelerates learning by adapting reward sensitivity to the policy's progress, as demonstrated by faster improvement on GL goals in \cref{fig:gl_training}.

\noindent\textbf{Task Unlocking.} Without progressive unlocking, complex tasks are trained before their prerequisite capabilities are sufficiently learned. Meanwhile, when simpler tasks gradually achieve higher success rates, success-weighted sampling reduces their training share, slowing their further improvement. Consequently, complex tasks must build on incompletely learned basic capabilities, leading to slower improvement. In contrast, our prerequisite-based unlocking allows the policy to first quickly acquire necessary simpler capabilities, and then achieve faster improvement on complex combinations once they are unlocked, as shown in \cref{fig:gl_training}. Moreover, training complex tasks on insufficiently learned capabilities may lead to suboptimal solutions that are difficult to overcome, potentially explaining the lower final success rates of W/o unlocking on type-change and full goals.

\noindent\textbf{Prerequisite Graph.} Removing the graph substantially reduces success on relatively more difficult type-change and full goals. Without task grouping, all goals share a kernel width that can shrink prematurely as simpler goals achieve higher success rates, leaving the harder type-change goals with insufficient reward guidance. Compared with W/o unlocking, which also trains all combinations from the start but retains task-wise kernel widths, type-change success drops further from 61.0\% to 38.6\%, highlighting the benefit of task-wise kernel annealing.

\noindent\textbf{Observation Encoding.} Replacing our nonlinear encoding with a linear one reduces success across all goal groups, especially for type-change and full goals with relatively larger goal distances. Moreover, it shows a larger performance gap between default and relaxed criteria than ours. This suggests that linear encoding struggles to support fine adjustments near the goal, particularly when the observations also span large goal distances. In contrast, our nonlinear encoding emphasizes small spatial deviations while compressing large ones, enabling more precise goal reaching across different manipulation scales.

\section{Conclusion}
\label{sec:conclusion}
We presented \method, an RL framework in which a single policy performs grasp transition with object relocation, given only the final target grasp and desired object world pose. A prerequisite graph over four basic capabilities lets the policy learn complex combinations through their simpler prerequisites, while a task-wise annealed kernel reward and a nonlinear observation encoding enable precise reaching across widely varying goal distances. Experiments show that one policy can generate diverse motions with different target grasps and object world poses, synthesize long-horizon motions with a sequence of discrete goals, and generalize to unseen everyday objects. The framework also applies to different dexterous hands.

{
    \small
    \bibliographystyle{ieeenat_fullname}
    \bibliography{main}
}

\setcounter{section}{0}
\setcounter{figure}{0}
\setcounter{table}{0}
\setcounter{equation}{0}
\renewcommand{\thesection}{S\arabic{section}}
\renewcommand{\thetable}{S\arabic{table}}
\renewcommand{\thefigure}{S\arabic{figure}}
\renewcommand{\theequation}{S\arabic{equation}}
\renewcommand{\theHsection}{S\arabic{section}}
\renewcommand{\theHtable}{S\arabic{table}}
\renewcommand{\theHfigure}{S\arabic{figure}}
\renewcommand{\theHequation}{S\arabic{equation}}
\maketitlesupplementary
\section{Training Details}
\label{sec:supp_training}

\Cref{tab:supp_randomization} lists the domain randomization applied during training for both hands. \Cref{tab:supp_params,tab:supp_pose_params} list the training parameters. Equation and section numbers in these tables refer to the main paper.

The kernel width $\sigma$ for each task is progressively reduced through the sequence 10, 8, 6.4, 5.12, 4.1, 3.28, 2.62, 2, 1.4, 1, 0.7, and 0.5, approximately following a multiplicative decay factor of 0.8 per step.

\begin{table}[h]
  \centering
  \caption{Domain Randomization in training.}
  \label{tab:supp_randomization}
  \footnotesize
  \begin{tabularx}{\columnwidth}{l|L}
    \toprule
    Quantity & Range \\
    \midrule
    Object mass & $[0.15, 0.30]$\,kg \\
    Friction of hand and object & $[0.5, 1.0]$ \\
    Center of mass on the long axis & $[0.25, 0.75]$ of the length \\
    \midrule
    Finger configuration noise & $[-0.02, 0.02]$\,rad \\
    Force pulse probability & $0.25$ \\
    Force pulse, Gaussian per axis & std.\ $2$\,N/kg $\times$ mass \\
    \midrule
    Finger PD stiffness & $[5.8, 6.2]$\,N\,m/rad \\
    Finger PD damping & $[0.127, 0.156]$\,N\,m\,s/rad \\
    \bottomrule
  \end{tabularx}
\end{table}

\begin{table}[h]
  \centering
  \caption{Position and rotation parameters, the same for both hands.}
  \label{tab:supp_pose_params}
  \footnotesize
  \newcommand{\grp}[1]{\midrule\multicolumn{3}{c}{\textit{#1}} \\ \midrule}
  \begin{tabularx}{\columnwidth}{l|>{\centering\arraybackslash}X>{\centering\arraybackslash}X}
    \toprule
    Parameter & Position & Rotation \\
    \grp{Goal distance, Sec.~3.3}
    Scales $s_p^X$, $s_r^X$ & 1\,cm & $10^\circ$ \\
    Threshold $\kappa$ of $\psi$ & 5 & 1.5 \\
    \grp{Observation encoding, Eq.~(5)}
    $d_0$, $d_{\max}$ of ${}_{h}\mathbf{e}_t^G$ & 1\,cm, 0.3\,m & 0.1\,rad, $\pi$\,rad \\
    $d_0$, $d_{\max}$ of ${}_{h}\mathbf{e}_t^C$ & 1\,mm, 0.3\,m & 0.01\,rad, $\pi$\,rad \\
    $d_0$, $d_{\max}$ of ${}_{h}\mathbf{e}_t^H$ & 1\,cm, 0.3\,m & 0.01\,rad, $\pi$\,rad \\
    $d_0$, $d_{\max}$ of ${}_{h}\mathbf{v}_t$ & 2\,mm, 0.15\,m & \\
    \grp{Action, Eq.~(6)}
    Scales $\alpha_p$, $\alpha_r$ & 3\,cm & 0.06\,rad \\
    \bottomrule
  \end{tabularx}
\end{table}

\begin{table}[t]
  \centering
  \caption{Training parameters for MANO, with Sharpa values where they differ.}
  \label{tab:supp_params}
  \footnotesize
  \newcommand{\grp}[1]{\midrule\multicolumn{2}{c}{\textit{#1}} \\ \midrule}
  \begin{tabularx}{\columnwidth}{l|L}
    \toprule
    Parameter & Value \\
    \grp{Simulation and training}
    Physics rate & 120\,Hz \\
    Control rate & 20\,Hz \\
    Parallel environments & 8{,}000 \\
    Episode length & 600 control steps \\
    Training time & 48\,h, one RTX 4090 \\
    \grp{PPO and network}
    Rollout length & 16 control steps \\
    Minibatches & 4 \\
    Epochs & 5 \\
    Learning rate & $3\times10^{-4}$, KL-adaptive \\
    Discount factor & 0.99 \\
    GAE parameter & 0.95 \\
    Clip range & 0.2 \\
    Entropy coefficient & 0 \\
    Separate actor and critic MLPs & 512-256-128, ELU \\
    \grp{Goal distance and reward, Secs.~3.3 and 3.4}
    Dead band $\delta_q$ of $P_t$ & $5^\circ$ \\
    Joint tolerance $s_q$ of $P_t$ & $10^\circ$ \\
    Kernel term weight & 1 \\
    Drop penalty $r_t^{\mathrm{drop}}$ & $-30$ \\
    Contact weight $\lambda_c$ & 0.075 \\
    Torque weight $\lambda_\tau$ & 0.01 \\
    Action smoothness weight $\lambda_a$ & $7\times10^{-4}$ \\
    \grp{Observation and action, Eqs.~(4) to (6)}
    Actor input size & 141, Sharpa 147 \\
    Critic input size & 294, Sharpa 302 \\
    Action size & 26, Sharpa 28 \\
    Action scale $\alpha_q$ & 0.0417\,rad \\
    \grp{Curriculum, Secs.~3.2 to 3.4}
    Success threshold $\eta$ & 0.4 \\
    Learned width $\sigma_{\mathrm{learned}}$ & 2 \\
    Success rate window & last 400 goals per width \\
    Completion in training & $E_t<\sigma$ for 5 steps \\
    Fixed tolerance of $p_i$ & $E_t<1$ \\
    \bottomrule
  \end{tabularx}
\end{table}

\section{Everyday Objects}
\label{sec:supp_everyday_objects}

\begin{table*}[t]
  \centering
  \caption{Everyday object dimensions (mm) and per-task success rates (\%) under the default criteria. Dashes indicate untested tasks.}
  \label{tab:supp_everyday_objects}
  \footnotesize
  \setlength{\tabcolsep}{7.5pt}
  \newcommand{\grp}[1]{\midrule\multicolumn{14}{c}{\textit{#1}} \\ \midrule}
  \resizebox{\textwidth}{!}{%
  \begin{tabular}{l|cccc|cccc|cccc|c}
    \toprule
    & \multicolumn{4}{c|}{Size} & \multicolumn{4}{c|}{Same type} & \multicolumn{4}{c|}{Type change} & \multicolumn{1}{c}{Full} \\
    \cmidrule(lr){2-5}\cmidrule(lr){6-9}\cmidrule(lr){10-13}\cmidrule(lr){14-14}
    Object & Length & Held length & Width & Thickness & TL & RL & RTL & Avg. & GL & TGL & RGL & Avg. & RTGL \\
    \grp{Everyday use needs a grasp type change}
    Short screwdriver & 190 & 114 & 40 & 37 & 99.0 & 90.3 & 98.4 & 95.9 & 77.0 & 71.6 & 68.8 & 72.5 & 63.7 \\
    Long screwdriver & 294 & 144 & 40 & 35 & 99.8 & 91.1 & 99.6 & 96.8 & 81.3 & 66.7 & 75.6 & 74.5 & 59.9 \\
    Flashlight & 228 & 228 & 46 & 40 & 96.0 & 85.5 & 96.4 & 92.6 & 58.5 & 46.8 & 52.0 & 52.4 & 43.3 \\
    Cucumber & 228 & 228 & 30 & 29 & 95.2 & 52.2 & 95.4 & 80.9 & 9.3 & 12.9 & 11.5 & 11.2 & 8.5 \\
    Spice jar & 109 & 109 & 48 & 42 & -- & 96.0 & -- & 96.0 & 85.1 & -- & 73.4 & 79.3 & -- \\
    Carrot & 226 & 196 & 29 & 27 & 96.8 & 82.5 & 98.4 & 92.6 & 69.2 & 56.2 & 59.5 & 61.6 & 55.6 \\
    Bottle & 140 & 100 & 46 & 36 & -- & 89.5 & -- & 89.5 & 75.6 & -- & 66.9 & 71.3 & -- \\
    Corn & 165 & 165 & 46 & 44 & 98.6 & 92.7 & 97.4 & 96.2 & 79.6 & 65.9 & 73.0 & 72.8 & 62.3 \\
    \grp{Everyday use needs no grasp type change}
    Mallet & 297 & 237 & 27 & 22 & 96.8 & 96.8 & 93.3 & 95.6 & -- & -- & -- & -- & -- \\
    Hammer & 195 & 155 & 23 & 18 & 84.9 & 85.5 & 81.9 & 84.1 & -- & -- & -- & -- & -- \\
    Spatula & 348 & 140 & 26 & 25 & 97.4 & 97.6 & 94.4 & 96.5 & -- & -- & -- & -- & -- \\
    Spoon & 278 & 160 & 28 & 27 & 98.8 & 99.0 & 97.8 & 98.5 & -- & -- & -- & -- & -- \\
    Marker & 158 & 158 & 28 & 28 & 96.0 & 95.4 & 94.8 & 95.4 & -- & -- & -- & -- & -- \\
    Pipe wrench & 405 & 222 & 30 & 17 & 87.9 & 78.8 & 73.0 & 79.9 & -- & -- & -- & -- & -- \\
    Wrench & 254 & 125 & 25 & 25 & 98.6 & 98.2 & 97.2 & 98.0 & -- & -- & -- & -- & -- \\
    Ratchet & 203 & 165 & 28 & 28 & 96.6 & 96.0 & 96.0 & 96.2 & -- & -- & -- & -- & -- \\
    Cream tube & 195 & 171 & 56 & 41 & 76.2 & 64.7 & 70.4 & 70.4 & -- & -- & -- & -- & -- \\
    Water bottle & 142 & 117 & 66 & 48 & 63.1 & 19.2 & 22.2 & 34.8 & -- & -- & -- & -- & -- \\
    Spray can & 127 & 116 & 50 & 49 & 99.6 & 99.4 & 99.6 & 99.5 & -- & -- & -- & -- & -- \\
    Banana & 230 & 185 & 48 & 43 & 89.3 & 92.9 & 88.3 & 90.2 & -- & -- & -- & -- & -- \\
    \midrule
    Avg. across objects & 220.6 & 161.8 & 38.0 & 33.1 & 92.8 & 85.2 & 88.6 & 88.9 & 67.0 & 53.3 & 60.1 & 60.1 & 48.9 \\
    \bottomrule
  \end{tabular}%
  }
  \vspace{5mm}
\end{table*}

\begin{figure*}[!ht]
  \centering
  \includegraphics[width=\textwidth]{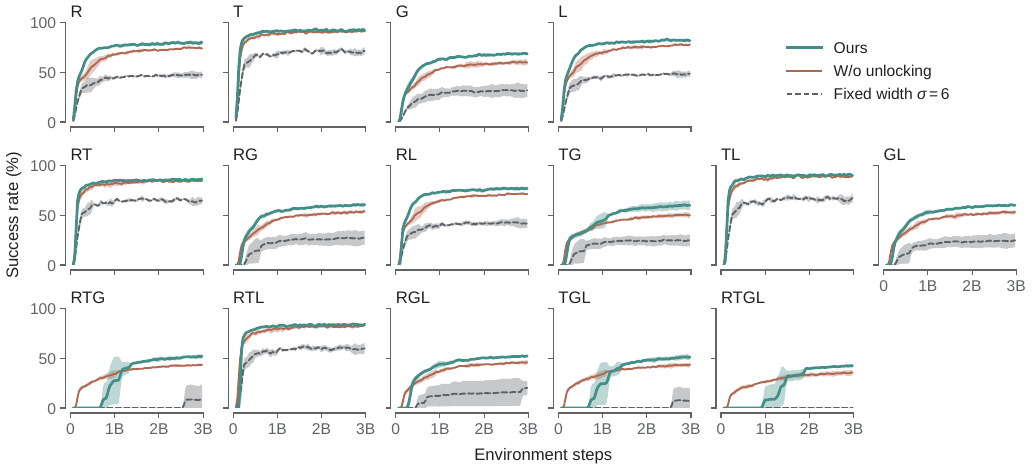}
  \caption{Training success rates of ours, W/o unlocking, and Fixed width $\sigma=6$ on all 15 tasks under the default criteria. Success is zero before task unlocking. Curves are smoothed over 500 policy updates. Each line shows the mean across three seeds, with the shaded area indicating the standard deviation.}
  \label{fig:supp_training_curves}
\end{figure*}

We evaluate the zero-shot generalization of our policy to everyday objects with geometries beyond the training distribution. The evaluation set comprises ten tools and ten household items, shown at a consistent scale in Fig.~6 of the main paper. We select objects that are typically held by a handle or an elongated body, using either a pinch grasp near one end or an enveloping grasp around the held part. These objects exhibit substantial geometric differences from the training objects, including flat handles, curved bodies, flattened tubes, and irregular or tapered surfaces. Five of the tools are adapted from DexToolBench~\cite{simtoolreal2026}. For each object, we construct a grasp library using the same grasping policies employed for the training objects, retaining only grasps on the intended holding region.

We assign each object a set of capabilities based on its typical everyday use and evaluate it only on goals involving the corresponding capabilities. All objects are evaluated on in-hand reorientation and object relocation. In-hand translation needs room for the hand to slide along the held part, so we evaluate it only when the held part is at least 110\,mm long. This excludes the spice jar and bottle. We further divide the objects into two groups based on whether their typical use requires a grasp type change. For example, a screwdriver is typically carried with an enveloping grasp but requires a pinch grasp to drive a screw, whereas hammers and wrenches are typically manipulated using only an enveloping grasp.

\Cref{tab:supp_everyday_objects} reports the dimensions of each object, the specific tasks evaluated, and their corresponding success rates under the default criteria. Length spans the whole object, while held length spans only the held part. Width and thickness are defined as the maximum and minimum extents of the median cross-section of the held part, respectively. The goals follow Sec.~4.1 of the main paper, with object relocation included in every goal. The last row reports the mean dimensions across all 20 objects and the average success rate for each task across all applicable objects, corresponding to the results in Tab.~2 of the main paper.

\makeatletter
\setlength{\@dblfptop}{0pt}
\setlength{\@dblfpsep}{\dblfloatsep}
\setlength{\@dblfpbot}{0pt plus 1fil}
\makeatother

\section{Training Curves on All Tasks}
\label{sec:supp_training_curves}

\Cref{fig:supp_training_curves} shows the training curves of all 15 tasks for ours, W/o unlocking, and Fixed width $\sigma=6$. Figure~7 of the main paper shows the same curves for GL goals, and Sec.~4.5 of the main paper describes the settings. Our method overall demonstrates higher final success rates, faster improvement once the task is unlocked, and smaller variance among different seeds.

\end{document}